# Enhanced high dynamic range 3D shape measurement based on generalized phase-shifting algorithm

Minmin Wang[a], Guangliang Du[a], Canlin Zhou*[a], Chaorui Zhang[a], Shuchun Si[a], Hui Li[a], Zhenkun Lei[b], YanJie Li[c]

(a. School of Physics, Shandong University, Jinan, 250100, China

b. Department of engineering mechanics, Dalian University of Technology, Dalian, 116024

c. School of civil engineering and architecture, Jinan University, Jinan, 2250020 )

*Corresponding author: Tel: +8613256153609;   E-mail address: canlinzhou@sdu.edu.cn;

**Corresponding author: Tel: +8615841175236;   E-mail address: leizk@163.com

## Abstract

It is a challenge for Phase Measurement Profilometry (PMP) to measure objects with a large range of reflectivity variation across the surface. Saturated or dark pixels in the deformed fringe patterns captured by the camera will lead to phase fluctuations and errors. Jiang et al. proposed a high dynamic range real-time 3D shape measurement method without changing camera exposures. Three inverted phase-shifted fringe patterns are used to complement three regular phase-shifted fringe patterns for phase retrieval when any of the regular fringe patterns are saturated. But Jiang's method still has some drawbacks: (1) The phases in saturated pixels are respectively estimated by different formulas for different cases. It is shortage of an universal formula; (2) it cannot be extended to four-step phase-shifting algorithm because inverted fringe patterns are the repetition of regular fringe patterns; (3) only three unsaturated intensity values at every pixel of fringe patterns are chosen for phase demodulation, lying idle the other unsaturated ones. We proposed a method for enhanced high dynamic range 3D shape measurement based on generalized phase-shifting algorithm, which combines the complementary technique of inverted and regular fringe patterns with generalized phase-shifting algorithm. Firstly, two sets of complementary phase-shifted fringe patterns, namely regular and inverted fringe patterns are projected and collected. Then all unsaturated intensity values at the same camera pixel from two sets of fringe patterns are selected, and employed to retrieve the phase by generalized phase-shifting algorithm. Finally, simulations and experiments are conducted to prove the validity of the proposed method. The results are analyzed and compared with Jiang's method, which demonstrate that the proposed method not only expands the scope of Jiang's method, but also improves the measurement accuracy.

## 1.Introduction

Phase measurement profilometry (PMP) based on the fringe projection is a well-developed technique and powerful tool in three-dimensional shape measurement for its superiorities of full-filed, high resolution and accuracy[1-5]. However, it is still a big challenging problem to measure objects with a large range of reflectivity variation across the surface. Owing to the limited dynamic range of projector and camera, saturated or dark regions will appear in the deformed fringe patterns, leading to low signal-noise ratio (SNR) and decreased accuracy even the failure of measurement.

Aiming to solve this problem, many solutions have been proposed. For example, Wolff[6] used polarizing plates to avoid the appearance of saturated pixels, but it reduces the output light intensity of the projector and the incoming light of the camera. Richard[7] avoided the reflection areas of mirror through multiangular measurement system, but it needs complex stitching process, limiting the measurement accuracy and speed. Zhao et al.[8] established a mathematical model of saturated pixels, using the intensities of unsaturated pixels to calculate the phase. The phase reconstruction algorithm proposed by Chen[9] decreased the phase deviation caused by partial intensity saturation, but this method is constrained by the saturated coefficients. Zhang et al.[10] put forward a high dynamic scanning technology by finding the maximum unsaturated strength, but it requires a subjective adjustment of camera exposures. Jiang et al.[11] proposed a 3-D scanning technique for high-reflective surfaces by introducing the HDRFA to phase-shifting method. Waddington et al.[12] presented a camera-independent method of avoiding image saturation using modified sinusoidal fringe-pattern projection, but it needs complex pre-calibration. The method presented by Babai et al.[13] recursively controlled the intensity of the projection pattern at pixel level based on the feedback of acquired fringe patterns. Zhao et al.[14] developed a fast and high dynamic range digital fringe projector based on DLP and intensity modulation principle. Two methods proposed by Ekstrand et al.[15-16] used defocusing binary projection to solve the problem that the camera exposure time cannot be arbitrarily chosen because DLP produces gray scale values by time modulation. One of the method is to determine the optimum exposure time while the other one is to combine multiple fringe images into composite HDR fringe images. Zhong et al.[17] presented an optimal exposure time calibration method based on the analysis of camera images under sinusoidal fringe illumination. Long et al. 's method[18] used the magnitude of a non principal frequency component to identify saturated pixels, then it was combined with high dynamic scanning[10] to measure high dynamic range objects. Lin et al.[19] proposed an adaptive projection technology, adaptively adjusting the pixel-wise intensities of the projected fringe patterns based on the saturated pixels in the captured images of the surface. Jeong et al.[20] used a spatial light modulator to prevent the image sensor from being saturated. Basel et al.[21] proposed a multi-polarization fringe projection (MPFP) imaging technique that eliminates saturated points and enhances the fringe contrast by selecting proper polarized channel measurements, but it needs expensive polarization cameras. Chen et al.[22] measured high dynamic range shapes directly using saturated fringes, but many fringe patterns are needed.

Recently, Jiang et al.[23] proposed a high dynamic range real-time 3D shape measurement method without changing camera exposures. Though the three-step algorithm is the simplest phase-shifting algorithm that requires the minimum number of fringe patterns. But Zuo et al.[24] found that four-step phase-shifting algorithm is less sensitive to the high order harmonics. To avoid the complex gamma correction[25], it's better to adopt more than three-step phase-shifting algorithm to obtain a satisfactory result.

According to our own experiences, Jiang's method has the following disadvantages: (1) The phases in saturated pixels are respectively estimated by different formulas for different cases. It is shortage of an universal formula; (2) it cannot be extended to four-step phase-shifting algorithm because inverted fringe patterns are the repetition of regular fringe patterns; (3) only three unsaturated fringe intensities are chosen for phase demodulation, lying idle the other unsaturated ones, e.g. Eq. (8) in Ref. 23 uses only inverted pattern $I_1^{inv}$ to replace regular pattern $I_1$ when $I_1$ is saturated, lying idle $I_2^{inv}$ and $I_3^{inv}$.

Here, we proposed a method for enhanced high dynamic range 3D shape measurement based on generalized phase-shifting algorithm, which combines the complementary technique of inverted and regular fringe patterns with generalized phase-shifting algorithm, and it's an extension of Jiang's method. We tested the proposed method on simulations and experiments, and compared our results with Jiang's method. The results have proved the validity of our method.

The remainder of this paper is as follows. Section 2 describes the theory. Section 3 discusses the simulation results. Section 4 shows experimental validation. Section 5 summarizes the full paper.

## 2. Theory

Phase-shifting algorithm[26,27] is an active research field in phase measurement profilometry. A serial of sinusoidal fringe images with constant phase-shifting amount are projected on the target surface. Camera synchronously captures the distorted fringe patterns carrying the information about the object's depth. In ideal cases, camera can record the light intensity properly. However, in actual situation, conventional fringe patterns captured by camera fail to recover the depth data from HDR objects due to pixels saturation or low SNR of fringe patterns, both of which will lead to phase errors and unreliable 3D results.

Jiang et al.[23] proposed a three-step phase-shifting algorithm for optically capturing high-contrast 3D surfaces without changing camera exposures, the core idea of which is using the 180-degree phase-shifted (or inverted) fringe patterns to complement regular fringe patterns. We discovered that Jiang's method still has some drawbacks. Aiming to solve these questions, we proposed a method for enhanced high dynamic range 3D shape measurement based on generalized phase-shifting algorithm. The rest of this section will explain the details about the proposed method.

Here, we briefly review the generalized phase shifting algorithm[28-29], which is used to demodulate the phase by the least-square method. Regular phase-shifted fringe patterns captured by the camera are mathematically described as:

$$I_k(x,y) = A(x,y) + B(x,y)\cos[\phi(x,y)+\delta_k] \tag{1}$$

Where $(x,y)$ are Cartesian coordinates, $A(x,y)$ is the background intensity, $B(x,y)$ is the intensity modulation amplitude, $\delta_k$ is the arbitrary phase step, $\phi(x,y)$ is the phase to be retrieved, k = 1, 2, 3...N, k is image index, N is the total number of images.

As for N-step generalized phase-shifting algorithm, there are N equations with $A(x,y)$, $B(x,y)$ and $\phi(x,y)$ as unknowns. Eq. (1) can be rewritten as,

$$I_k(x,y) = A(x,y) + B(x,y)\cos[\phi(x,y)]\cos(\delta_k) - B(x,y)\sin[\phi(x,y)]\sin(\delta_k) \quad (2)$$

Assuming that $p(x,y) = B(x,y)\cos[\phi(x,y)]$, $q(x,y) = B(x,y)\sin[\phi(x,y)]$, Eq. (2) can be described as,

$$I_k(x,y) = A(x,y) + p(x,y)\cos(\delta_k) - q(x,y)\sin(\delta_k) \quad (3)$$

The deviation square sum for all N fringe patterns can be expressed as,

$$E(x,y) = \sum_{k=1}^{N} \{A(x,y) + p(x,y)\cos(\delta_k) - q(x,y)\sin(\delta_k) - I'_k(x,y)\}^2 \quad (4)$$

Where $I'_k$ is the intensity of fringe patterns measured in the experiment. According to the principle of the least-square method, the extreme value condition of Eq. (4) should be satisfied, which can be expressed as,

$$\frac{\partial E(x,y)}{\partial p(x,y)} = 0, \quad \frac{\partial E(x,y)}{\partial q(x,y)} = 0, \quad \frac{\partial E(x,y)}{\partial A(x,y)} = 0 \quad (5)$$

In this way, three unknowns of p(x,y), q(x,y) and A(x,y) can be resolved simultaneously from solving these equations. Solve Eq. (5) to get the following matrices,

$$X = U^{-1}Q \quad (6)$$

Where

$$U = \begin{bmatrix} \sum_{k=1}^{N}\cos^2(\delta_k) & -\sum_{k=1}^{N}\sin(\delta_k)\cos(\delta_k) & \sum_{k=1}^{N}\cos(\delta_k) \\ -\sum_{k=1}^{N}\sin(\delta_k)\cos(\delta_k) & \sum_{k=1}^{N}\sin^2(\delta_k) & -\sum_{k=1}^{N}\sin(\delta_k) \\ \sum_{k=1}^{N}\cos(\delta_k) & -\sum_{k=1}^{N}\sin(\delta_k) & N \end{bmatrix}$$

$$X = [p(x,y) \quad q(x,y) \quad A(x,y)]^T$$

$$Q = [\sum_{k=1}^{N}\cos(\delta_k)I_k(x,y) \quad -\sum_{k=1}^{N}\sin(\delta_k)I_k(x,y) \quad \sum_{k=1}^{N}I_k(x,y)]^T \quad (7)$$

Assume that $a = \sum_{k=1}^{N}\cos^2(\delta_k)$, $b = \sum_{k=1}^{N}\sin^2(\delta_k)$, $c = \sum_{k=1}^{N}\cos(\delta_k)$, $d = \sum_{k=1}^{N}\sin(\delta_k)$,

$$e = \sum_{k=1}^{N} \sin(\delta_k)\cos(\delta_k), \quad f = \sum_{k=1}^{N} \sin(\delta_k) I_k(x,y), \quad g = \sum_{k=1}^{N} \cos(\delta_k) I_k(x,y), \quad h = \sum_{k=1}^{N} I_k(x,y)$$

Solving Eq. (6) leads to,

$$q(x,y) = (Ng - hc)(Ne - cd) - (Nf - hd)(Na - c^2) \quad (8)$$

$$p(x,y) = (Ng - hc)(Nb - d^2) - (Nf - hd)(Ne - dc) \quad (9)$$

Then the wrapped phase of the measured object can be obtained by,

$$\phi(x,y) = \arctan[\frac{q(x,y)}{p(x,y)}] = \arctan[\frac{(Ng - hc)(Ne - cd) - (Nf - hd)(Na - c^2)}{(Ng - hc)(Nb - d^2) - (Nf - hd)(Ne - dc)}] \quad (10)$$

Here, we use the idea in Ref. 23, namely projecting and acquiring two sets of complementary phase-shifting sinusoidal fringe patterns respectively. The inverted fringe patterns corresponding to the regular fringe patterns in Eq. (1) are expressed as,

$$I_k^{inv}(x,y) = A(x,y) + B(x,y)\cos[\phi(x,y) + \delta_k + \pi] \quad (11)$$

Thus, a total of 2N pictures are collected from regular and inverted fringe patterns, and at least three intensities at the same pixel is needed to calculate the corresponding phase. As the phase-step is known, for saturated pixels, all unsaturated intensities from 2N frame fringe patterns can be picked out and used to estimate the phase by the generalized phase-shifting algorithm. Therefore, for every pixel, if the intensity values of m frame patterns are saturated, the remaining 2N-m fringe patterns will be selected and processed by 2N-m step generalized phase-shifting algorithm. The same procedure can be performed pixel by pixel and the whole phase map can be evaluated.

The main stages of our algorithm are summarized as follows：

(1) design the regular and inverted fringe patterns according to Eq. (1) and (11), and project them onto the object's surface;

(2) capture the deformed fringe patterns by a CCD camera;

(3) determine the phase distribution pixel by pixel:

The intensity values (i.e. $I_1$, $I_2$, $\cdots$, $I_N$, $I_1^{inv}$, $I_2^{inv}$, $\cdots$, $I_N^{inv}$) of 2N frame images at the same pixel position are obtained from regular and inverted fringe patterns. Saturated intensity values are removed, and the remaining all unsaturated intensity values are employed to determine the phase by generalized phase-shifting algorithm. The same procedure can be performed pixel by pixel. Thus the whole phase distribution can be obtained;

(4) unwrap the wrapped phase by the unwrapping algorithm;

(5) transform the absolute phase to 3D height information after the system is calibrated.

## 3. Numerical simulation

In order to verify the feasibility of the proposed method, numerical simulations were carried out. We simulated different amount of saturation by a scale factor $S$[21-22]. For the regular fringe patterns, the *k*-th ideal 8-bit fringe patterns for a N-step generalized phase-shifting algorithm can be described as,

$$I_k(x,y) = S \times 127.5[1+\cos(2\pi x/p + \phi(x,y) + \delta_k)] \tag{12}$$

Where $p$ is the period of fringe patterns, $\delta_k$ is the arbitrary phase-step, $\phi(x,y)$ is the phase to be retrieved, and the *Peaks* function of MATLAB was used to represent the phase corresponding to the measured object. Its shape is shown in Fig. 1, of which the resolution is 512×512 pixels. Assume that an 8-bit camera is used, if $S > 1$, fringe patterns are saturated and they can be mathematically described as,

$$I_k''(x,y) = \begin{cases} I_k(x,y) & I_k(x,y) \leq 255 \\ 255 & otherwise \end{cases} \tag{13}$$

For the inverted fringe patterns, as counterparts of the regular fringe patterns in Eq. (12), they can be written as,

$$I_k^{inv}(x,y) = S \times 127.5[1+\cos(2\pi x/p + \varphi(x,y) + \pi + \delta_k)] \tag{14}$$

Their saturation can be simulated in the same manner as Eq. (13).

When Jiang's method was extend to the four-step phase-shifting algorithm with equal phase steps of π/2, it failed to recover the phase properly. According to Jiang's method[23], inverted fringe patterns are used to complement regular fringe patterns when any pixels of the regular fringe patterns are saturated. However, as the equal phase steps of π/2 bring about the repetition of inverted and regular fringe patterns, when the number of saturated regular patterns is greater than two, the amount of available intensities at the same pixel will be less than three, leading to the failure of the phase demodulation. Therefore, Jiang's method is not applicable when it is extend to the four-step algorithm with π/2 steps.

However, due to that the phase steps can be arbitrary in our proposed method, we can design them skillfully to ensure that there is no repetition of inverted and regular fringe patterns, therefore we can extend Jiang's method to the four-step phase-shifting algorithm.

Taking $N=4$ in Eq. (10), the four-step generalized phase-shifting formula for phase demodulation is as follows,

$$\phi(x,y) = \arctan[\frac{(4g-hc)(4e-cd)-(4f-hd)(4a-c^2)}{(4g-hc)(4b-d^2)-(4f-hd)(4e-dc)}] \tag{15}$$

We chose $\delta_1=5\pi/2$, $\delta_2=-\pi/6$, $\delta_3=5\pi/4$ and $\delta_4=-4\pi/5$. $S$ was set to 1.5. Thus eight frame phase-shifted fringe images were produced according to Eq. (12), (13) and Eq. (14), and one of those fringe patterns is shown in Fig. 2(a). Firstly, ignoring pixel saturation, directly use four frame regular fringe patterns to estimate the phase by Eq. (15). The result is shown in Fig. 2(b). The corresponding unwrapped phase includes many fluctuations, especially in the saturated areas. Then we adopted our method. For every pixel, all unsaturated intensity values were picked out from eight fringe patterns and employed to determine the phase according to Eq. (10). The result is presented as Fig. 2(c). Apparently, the phase fluctuations are greatly decreased. In addition, to present the accuracy of the proposed method, we give the residual errors between the theoretical value (Fig. 1) and the phase respectively obtained by conventional method and proposed method (Figs. 2(b) and 2(c)). Fig. 2(d) and 2(e) respectively show the residual errors along middle row. It is found that the maximum residual error are 0.4 rad and $1.5 \times 10^{-14}$ rad respectively. Clearly, this result demonstrates that the accuracy of phase retrieval with the proposed method is higher than the conventional method. The previous simulation did not include noise effect, and thus are difficult to be realized in practical implementations. The following simulations tested the robustness of the proposed method to noise. White noise with a variance of 20 and a mean of 200 was added to the fringe patterns. The process procedures were similar to the previous simulation. One of the deformed fringe patterns with white noise is presented in Fig. 3(a). And Fig. 3(b)-(c) respectively show the results obtained by conventional method and proposed method. Figs. 3(d) and 3(e) show the residual errors between the theoretical value (Fig.1) and Fig .3(b), 3 (c) along middle row. The results demonstrate that the proposed method is more robust to noise while the result by conventional method has more obvious ripples and its maximum residual error increases to 0.4 rad.

Subsequently, we discussed the influence of saturation level (i.e., scale factor $S$) on the performance of the proposed method. In order to show the differences between conventional method, Jiang's method and our method, we simulated different level of saturation using three-step phase-shifting algorithm with $2\pi/3$ step. The root mean square errors (RMSE) between the theoretical value and the retrieved phase are shown in Table 1. We can see that the RMSE of the proposed method or other methods are increased with higher saturation level. And the results by conventional method have larger RMSE. The differences of proposed method and Jiang's method are not significant, and their saturation tolerance is better than Jiang's method.

Table. 1. RMSE (rad) of phase retrieval with different method at different level of saturation

| | S | 1.0 | 1.2 | 1.4 | 1.6 | 1.8 | 2.0 | 2.2 |
|---|---|---|---|---|---|---|---|---|
| Signal with noise | Conventional method | 0.5611 | 2.3750 | 4.5230 | 6.3033 | 7.7112 | 8.8438 | 9.7811 |
| | Jiang's method | $3.3592\times 10^{-14}$ | $7.6250\times 10^{-14}$ | $7.9993\times 10^{-14}$ | $9.1093\times 10^{-14}$ | 0.0025 | 0.0377 | 0.2271 |
| | Proposed method | $3.3003\times 10^{-14}$ | $7.5461\times 10^{-14}$ | $7.9172\times 10^{-14}$ | $9.0145\times 10^{-14}$ | 0.0015 | 0.0238 | 0.1764 |

What's more, Eq. (14) in Ref. 23 is wrong, its correct formula should be described as,

$$\phi(x,y) = \tan^{-1}\{\frac{2[B_1(x,y) - B_3(x,y)]}{\sqrt{3}[B_2(x,y) - B_1(x,y) - B_3(x,y)]}\} \qquad (16)$$

## 4. experiment and result analysis

To verify the proposed algorithm, we developed a fringe projection measurement system, which consists of a DLP projector (Optoma DN344 ) and a CCD camera (DH-SV401FM). The camera uses a 25 mm focal length Mega-pixel lens (ComputarFAM2514-MP2), of which the resolution is 780 × 582 pixels, with a maximum frame rate of 50 frames / sec. The 3D measurement software is programmed by MATLAB.

Phase-shifted regular and inverted fringe patterns were projected onto the target by the DLP controlled by computer. The deformed fringe patterns were captured and saved by the camera, which can be analyzed to get the 3D surface shape, as described in section 2. The quality guided path unwrapping algorithm[30] was used to calculate the continuous phase map.

The test sample is a pair of scissors with high dynamic range of surface reflectivity variations, of which the photograph is shown in Fig. 4(a). Its handles are black and dark, requiring a high exposure for accurate measurement. But its blades are made of reflective metal, high reflection will lead to pixel saturature. Therefore, it is difficult for conventional phase-shifting algorithm to accurately measure the whole surface with a single exposure.

For conventional three-step phase-shifting algorithm, one of the low exposure fringe patterns without any saturated point is shown in Fig. 4(b). In this case, only bright metal areas are properly measured, but dark handle areas have very large amounts of noise due to the low SNR of fringe patterns. Fig. 4(e) shows such result. Besides, phase map has some ripples because of the gamma effect of the projector. As we increased the camera exposures, handle areas become more visible while pixels corresponding to bright metal areas become saturated, as shown in Fig. 4(c). In this case, dark areas are properly exposed, and the handle areas can be properly measured. However, the bright metal areas have large errors, as shown in Fig. 4(f). Imaging quality of dark handle areas is better under high exposure than that under low exposure. Besides, nonlinearity of projector produces some conspicuous ripples, and the corresponding phases of the saturated pixels in metal areas appear larger amounts of noise. So conventional three-step phase-shifting algorithm can not retrieve the phase accurately whether under low exposure or under high exposure.

However, if we projected and captured another set of inverted high exposure fringe patterns as Jiang did[23] (Fig. 4(d) shows one of the inverted, high exposure phase-shifted patterns corresponding to Fig. 4(c)), then processed the regular and inverted fringe patterns by Jiang's method, that is to say, for saturated pixels, the inverted fringe patterns would be used in lieu of the regular fringe patterns for phase computation. We can recover the phase as shown in Fig. 4(f). Obviously, it generates a better phase map, in which the error caused by saturated pixels in the bright metal areas is hardly noticeable, yet the gamma effect error is still there. We then applied our method to these deformed patterns. At every pixel position, all unsaturated fringe intensities from regular and inverted fringe patterns were selected and constructed more than three steps phase-shifting to retrieve the phase by generalized phase-shifting algorithm. In this way, ripples caused by the nonlinearity of the projector and camera are greatly reduced. The phase map of the scissors is faithfully reconstructed as shown in Fig. 4(h).

To better visualize the differences among these algorithms, we zoomed in the local blade part. Fig. 4(i), Fig. 4(j) and Fig. 4(k) are the close-up views for Fig. 4(f), Fig. 4(g) and Fig. 4(h) respectively. Clearly, for high exposure, conventional three-step phase-shifting algorithm brings about not only large noise in the bright metal areas because of the saturated pixels, but also obvious fluctuations in the phase map due to gamma effect. In contrast, Jiang's method has the capacity to eliminate the saturated error, but the phase fluctuations are still obvious due to the nonlinearity of the projector and camera. Our method can simultaneously reduce these errors. The comparison of the experimental results clearly demonstrates that the proposed method has a greater capability than the other methods to accurately measure the object with high-contrast surface.

## 5. Conclusion

In this paper, we proposed a method for enhanced high dynamic range 3D shape measurement based on generalized phase-shifting algorithm. It's an extension of Jiang's method, making Jiang's method applicable for four-step phase-shifting by designing the phase step skillfully. And it can simultaneously reduce both errors caused by saturated pixels and gamma effect of the system as all unsaturated fringe intensities at the same camera pixel from two sets of regular and inverted fringe patterns are selected, and employed to retrieve the phase by generalized phase-shifting algorithm. Simulations and experiments are carried out to demonstrate that the proposed method can not only expands the scope of Jiang's method, but also improves the measurement accuracy. The research result is valuable to engineering application in 3D shape measurement.

**Acknowledgment**

This work was supported by the National Natural Science Foundation of China (Grant nos. 11302082 and 11472070). The support is gratefully acknowledged. We would like to express our gratitude to Professor Zhang Song and Dr. Jiang Chufan for their invaluable help and useful discuss.

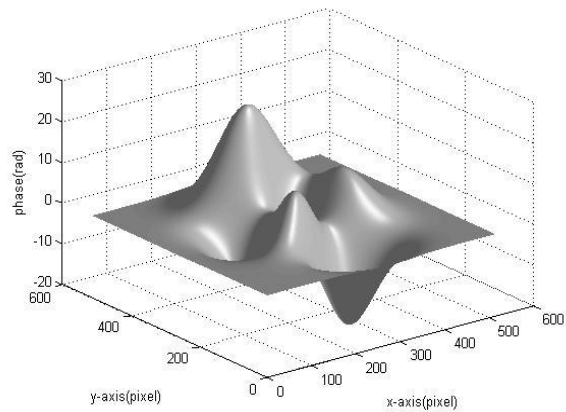

Fig. 1    phase distribution of peak function

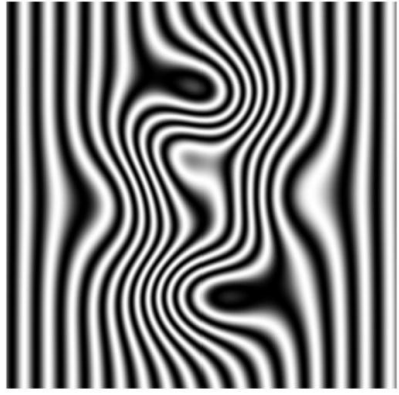

Fig. 2(a) one of the deformed fringe patterns

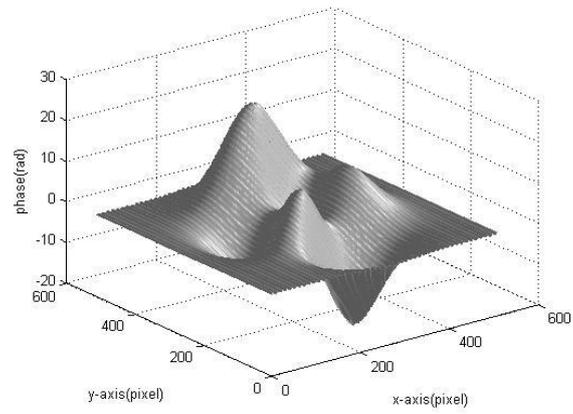

Fig. 2(b) phase obtained by conventional method

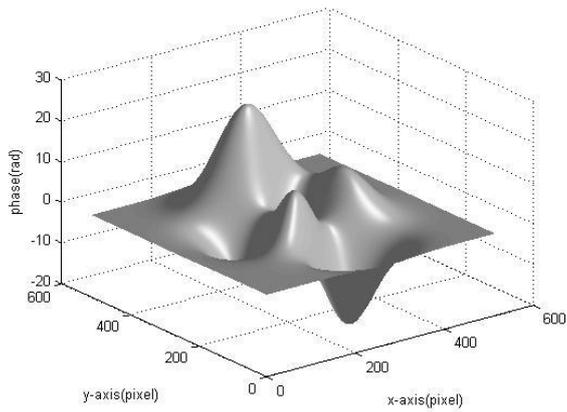

Fig. 2(c) phase obtained by the proposed method

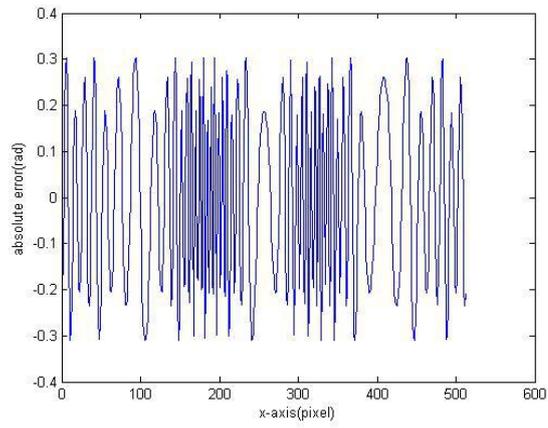

Fig. 2(d) the residual error between Fig. 2(b) and Fig 1 along middle row

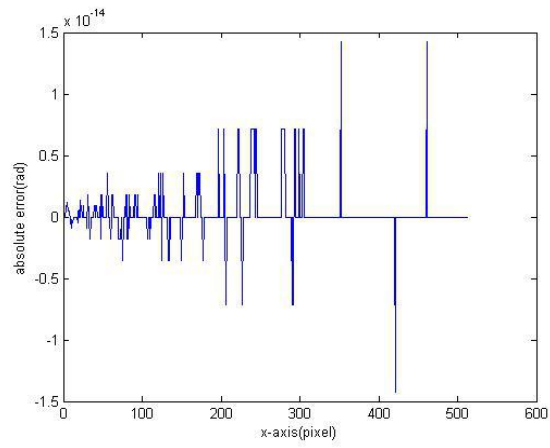

Fig. 2(e) the residual error between Fig. 2(c) and Fig 1 along middle row

Fig. 2 simulation results processed by 4-step phase-shifting algorithm

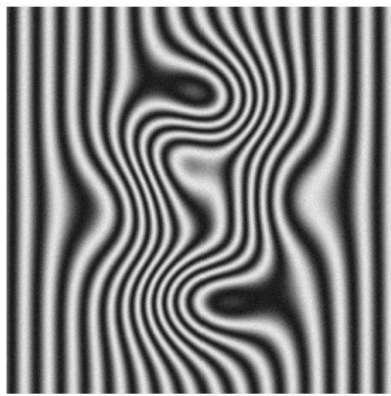

Fig. 3(a) one of the deformed fringe patterns with white noise

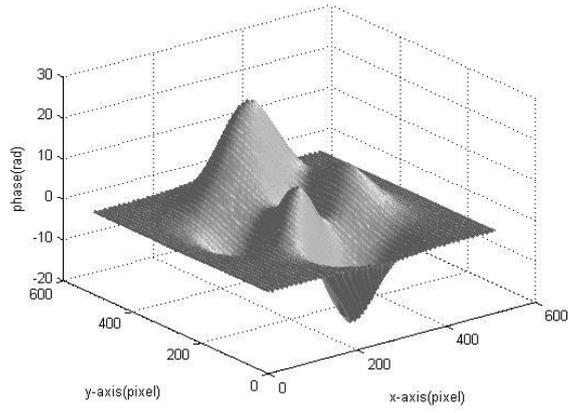

Fig. 3(b) phase obtained by conventional method

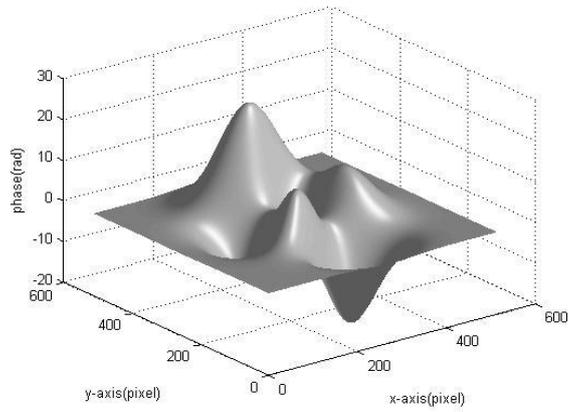

Fig. 3(c) phase obtained by the proposed method

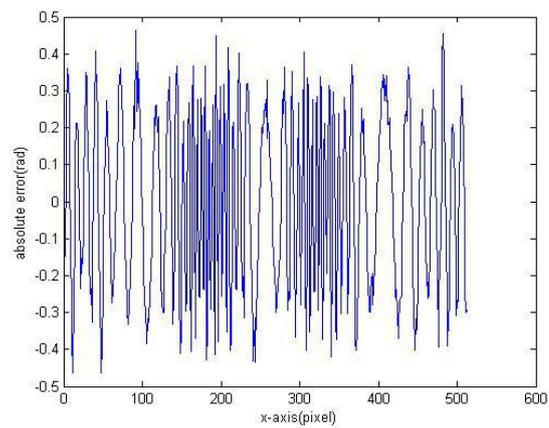

Fig. 3(d) the residual error between Fig. 3(b) and Fig 1 along middle row

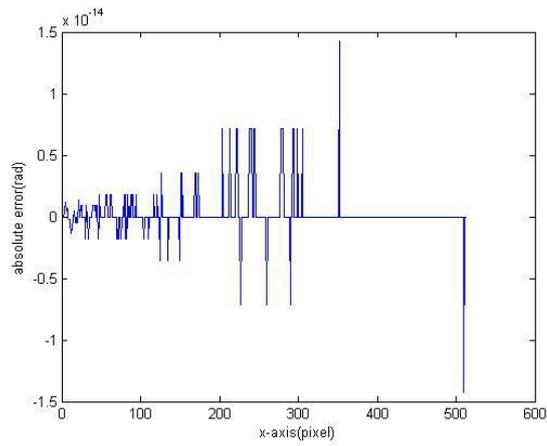

Fig. 3(e) the residual error between Fig. 3 (c) and Fig 1 along middle row

Fig. 3 simulation results obtained by 4-step phase-shifting method with white noise

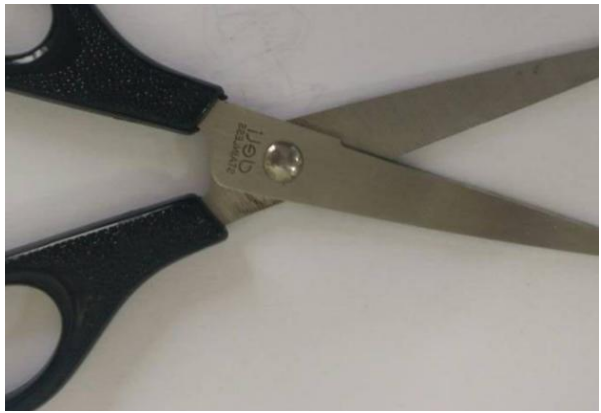

Fig. 4(a) the photograph of tested scissors

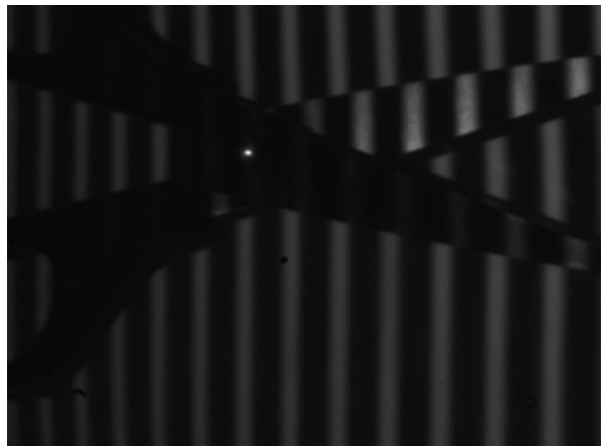

Fig. 4(b) one of the regular, low exposure phase-shifted patterns

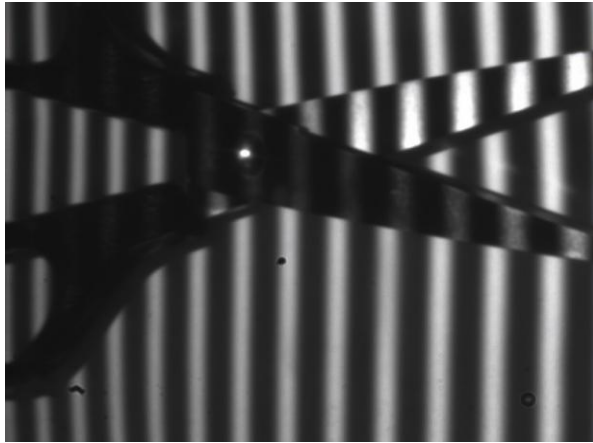

Fig. 4(c) one of the regular, high exposure phase-shifted patterns

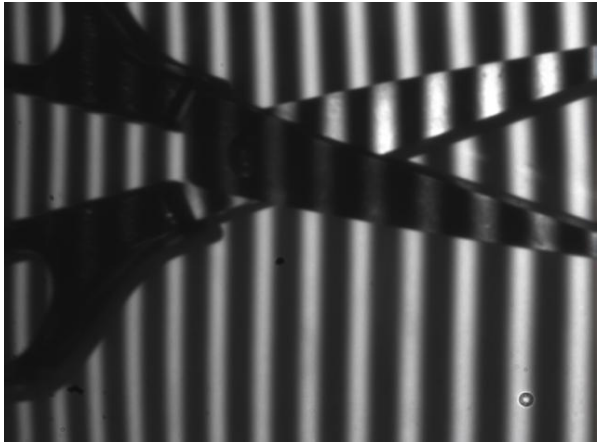

Fig. 4(d) one of the inverted, high exposure phase-shifted patterns corresponding to Fig. 4(c)

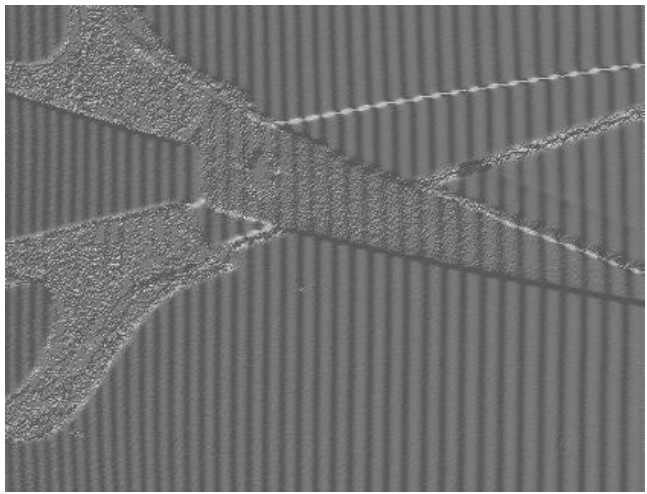

Fig. 4(e) phase map by conventional 3-step phase-shifting method from the low exposure patterns

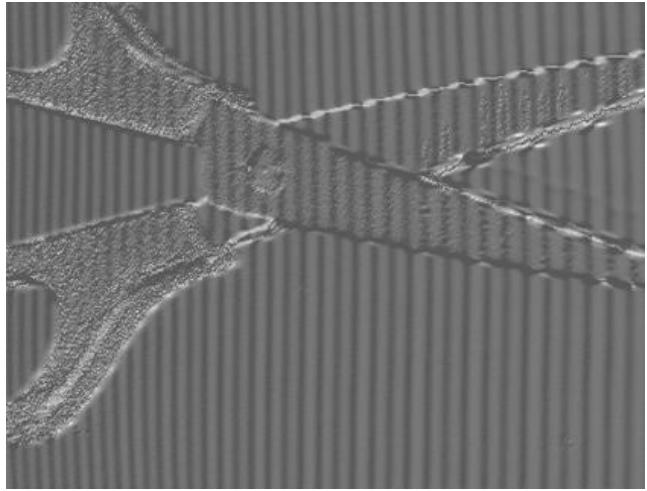

Fig. 4(f) phase map by conventional 3-step phase-shifting method from the high exposure patterns

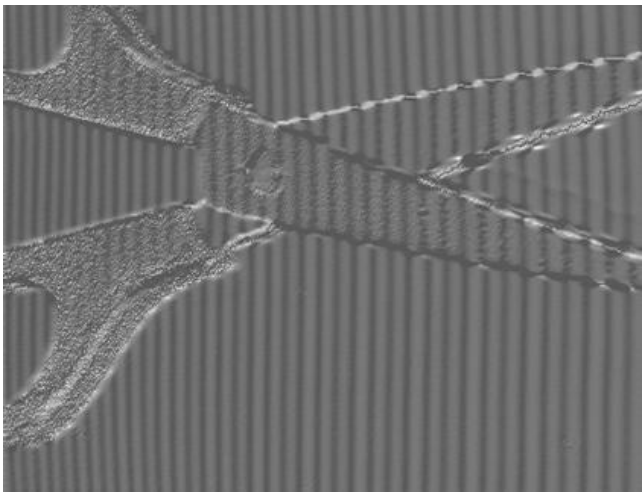

Fig. 4(g) phase map by Jiang's method from the high exposure patterns

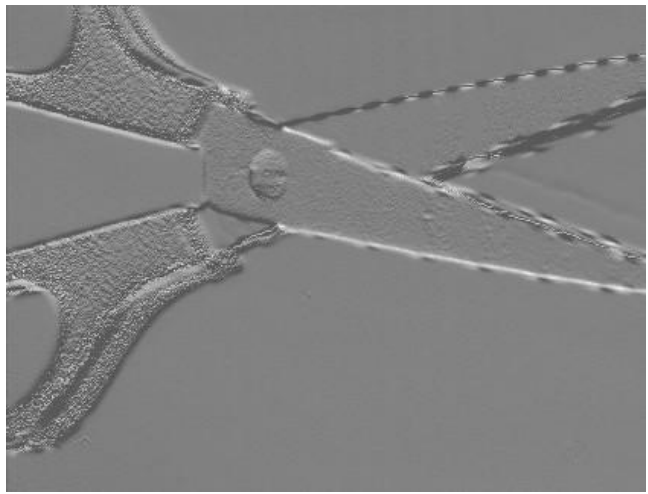

Fig. 4(h) phase map by the proposed algorithm from the high exposure patterns

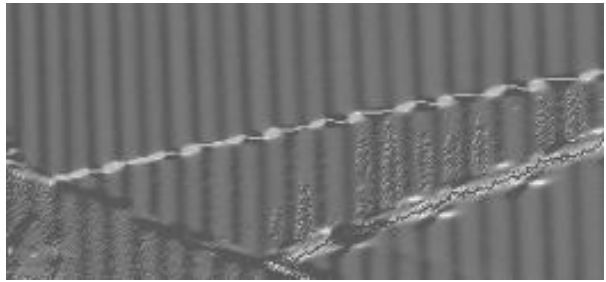

Fig. 4(i) zoomed-in result of blade area in Fig. 4(f)

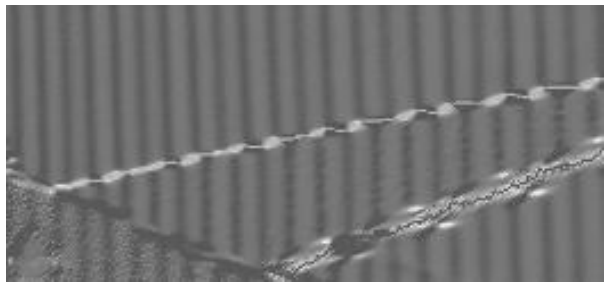

Fig. 4(j) zoomed-in result of blade area in Fig. 4(g)

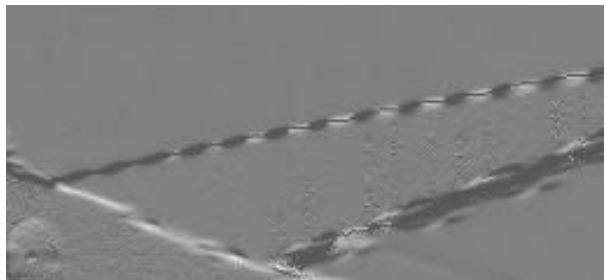

Fig. 4(k) zoomed-in result of blade area in Fig. 4(h)

Fig. 4 experimental results obtained by 3-step phase-shifting method